\documentclass{article}

\usepackage[nonatbib,final]{nips_2017}

\usepackage[utf8]{inputenc} 
\usepackage[T1]{fontenc}    
\usepackage{hyperref}       
\usepackage{url}            
\usepackage{booktabs}       
\usepackage{amsfonts}       
\usepackage{nicefrac}       
\usepackage{microtype}      

\usepackage{amsfonts}
\usepackage{color}
\usepackage{caption}
\usepackage{comment}
\usepackage{graphicx}
\usepackage{amsmath}
\usepackage{amssymb}
\usepackage{multirow}
\usepackage{floatrow}
\usepackage{svg}
\usepackage{enumitem}

\newcommand{\p}{\mathbf{p}}

\newcommand{\x}{\mathbf{x}}

\newcommand{\Z}{\mathbf{Z}}
\newcommand{\R}{\mathbf{R}}
\renewcommand{\t}{\mathbf{t}}

\newcommand{\K}{\mathbf{K}}
\renewcommand{\L}{\mathcal{L}}

\newcommand{\M}{\mathbf{M}}
\newcommand{\etal}{\emph{et al.}}
\newcommand{\ie}{\emph{i.e.} }
\newcommand{\eg}{\emph{e.g.} }

\newfloatcommand{capbtabbox}{table}[][\FBwidth]

\title{Learning Efficient Point Cloud Generation \\ for Dense 3D Object Reconstruction}

\author{
  Chen-Hsuan Lin, \; Chen Kong, \; Simon Lucey \\
  The Robotics Institute \\
  Carnegie Mellon University \\
  \texttt{chlin@cmu.edu}, \; \texttt{\{chenk,slucey\}@cs.cmu.edu}
}

\begin{document}

\maketitle

\begin{abstract}
Conventional methods of 3D object generative modeling learn volumetric predictions using deep networks with 3D convolutional operations, which are direct analogies to classical 2D ones.
However, these methods are computationally wasteful in attempt to predict 3D shapes, where information is rich only on the surfaces.
In this paper, we propose a novel 3D generative modeling framework to efficiently generate object shapes in the form of dense point clouds.
We use 2D convolutional operations to predict the 3D structure from multiple viewpoints and jointly apply geometric reasoning with 2D projection optimization.
We introduce the pseudo-renderer, a differentiable module to approximate the true rendering operation, to synthesize novel depth maps for optimization.
Experimental results for single-image 3D object reconstruction tasks show that we outperforms state-of-the-art methods in terms of shape similarity and prediction density.

\end{abstract}

\section{Introduction}

Generative models using convolutional neural networks (ConvNets) have achieved state of the art in image/object generation problems.  
Notable works of the class include variational autoencoders~\cite{kingma2013auto} and generative adversarial networks~\cite{goodfellow2014generative}, both of which have drawn large success in various applications~\cite{isola2016image,radford2015unsupervised,zhu2016generative,wang2016generative,yan2016attribute2image}.
With the recent introduction of large publicly available 3D model repositories~\cite{wu20153d,chang2015shapenet}, the study of generative modeling on 3D data using similar frameworks has also become of increasing interest.

In computer vision and graphics, 3D object models can take on various forms of representations.
Of such, triangular meshes and point clouds are popular for their vectorized (and thus scalable) data representations as well as their compact encoding of shape information, optionally embedded with texture.
However, this efficient representation comes with an inherent drawback as the dimensionality per 3D shape sample can vary, making the application of learning methods problematic.
Furthermore, such data representations do not elegantly fit within conventional ConvNets as Euclidean convolutional operations cannot be directly applied.
Hitherto, most existing works on 3D model generation resort to volumetric representations, allowing 3D Euclidean convolution to operate on regular discretized voxel grids.
3D ConvNets (as opposed to the classical 2D form) have been applied successfully to 3D volumetric representations for both discriminative~\cite{wu20153d,maturana2015voxnet,hegde2016fusionnet} and generative~\cite{girdhar2016learning,choy20163d,yan2016perspective,wu2016learning} problems.

Despite their recent success, 3D ConvNets suffer from an inherent drawback when modeling shapes with volumetric representations.
Unlike 2D images, where every pixel contains meaningful spatial and texture information, volumetric representations are information-sparse.
More specifically, a 3D object is expressed as a voxel-wise occupancy grid, where voxels ``outside'' the object (set to off) and ``inside'' the object (set to on) are unimportant and fundamentally not of particular interest.
In other words, the richest information of shape representations lies on the surface of the 3D object, which makes up only a slight fraction of all voxels in an occupancy grid.
Consequently, 3D ConvNets are extremely wasteful in both computation and memory in trying to predict much unuseful data with high-complexity 3D convolutions, severely limiting the granularity of the 3D volumetric shapes that can be modeled even on high-end GPU-nodes commonly used in deep learning research. 

In this paper, we propose an efficient framework to represent and generate 3D object shapes with dense point clouds.
We achieve this by learning to predict the 3D structures from multiple viewpoints, which is jointly optimized through 3D geometric reasoning.
In contrast to prior art that adopts 3D ConvNets to operate on volumetric data, we leverage \textit{2D convolutional operations} to predict points clouds that shape the \textit{surface} of the 3D objects.
Our experimental results show that we generate much denser and more accurate shapes than state-of-the-art 3D prediction methods.

Our contributions are summarized as follows:
\begin{itemize}[leftmargin=24pt]
\item We advocate that 2D ConvNets are capable of generating dense point clouds that shapes the surface of 3D objects in an undiscretized 3D space.
\item We introduce a pseudo-rendering pipeline to serve as a differentiable approximation of true rendering.
We further utilize the pseudo-rendered depth images for 2D projection optimization for learning dense 3D shapes.
\item We demonstrate the efficacy of our method on single-image 3D reconstruction problems, which significantly outperforms state-of-the-art methods.
\end{itemize}


\section{Related Work}

\paragraph{3D shape generation.}
As 2D ConvNets have demonstrated huge success on a myriad of image generation problems, most works on 3D shape generation follow the analogue using 3D ConvNets to generate volumetric shapes.
Prior works include using 3D autoencoders~\cite{girdhar2016learning} and recurrent networks~\cite{choy20163d} to learn a latent representation for volumetric data generation.
Similar applications include the use of an additional encoded pose embedding to learn shape deformations~\cite{yumer2016learning} and using adversarial training to learn more realistic shape generation~\cite{wu2016learning,gadelha20163d}.
Learning volumetric predictions from 2D projected observations has also been explored~\cite{yan2016perspective,rezende2016unsupervised,gadelha20163d}, which use 3D differentiable sampling on voxel grids for spatial transformations~\cite{jaderberg2015spatial}.
Constraining the ray consistency of 2D observations have also been suggested very recently~\cite{tulsiani2017multi}.

Most of the above approaches utilize 3D convolutional operations, which is computationally expensive and allows only coarse 3D voxel resolution.
The lack of granularity from such volumetric generation has been an open problem following these works.
Riegler~\etal~\cite{riegler2016octnet} proposed to tackle the problem by using adaptive hierarchical octary trees on voxel grids to encourage encoding more informative parts of 3D shapes.
Concurrent works follow to use similar concepts~\cite{hane2017hierarchical,tatarchenko2017octree} to predict 3D volumetric data with higher granularity.

Recently, Fan~\etal~\cite{fan2016point} also sought to generate unordered point clouds by using variants of multi-layer perceptrons to predict multiple 3D coordinates.
However, the required learnable parameters linearly proportional to the number of 3D point predictions and does not scale well; in addition, using 3D distance metrics as optimization criteria is intractable for large number of points.
In contrast, we leverage convolutional operations with a joint 2D project criterion to capture the correlation between generated point clouds and optimize in a more computationally tractable fashion.

\paragraph{3D view synthesis.}
Research has also been done in learning to synthesize novel 3D views of 2D objects in images.
Most approaches using ConvNets follow the convention of an encoder-decoder framework.
This has been explored by mixing 3D pose information into the latent embedding vector for the synthesis decoder~\cite{tatarchenko2016multi,zhou2016view,park2017transformation}.
A portion of these works also discussed the problem of disentangling the 3D pose representation from object identity information~\cite{kulkarni2015deep,yang2015weakly,reed2015deep}, allowing further control on the identity representation space.

The drawback of these approaches is their inefficiency in representing 3D geometry --- as we later show in the experiments, one should explicitly factorize the underlying 3D geometry instead of implicitly encoding it into mixed representations.
Resolving the geometry has been proven more efficient than tolerating in several works (\eg Spatial Transformer Networks~\cite{jaderberg2015spatial,lin2016inverse}).

\section{Approach}

\begin{figure*}[t!] \center
\includegraphics[width=\linewidth]{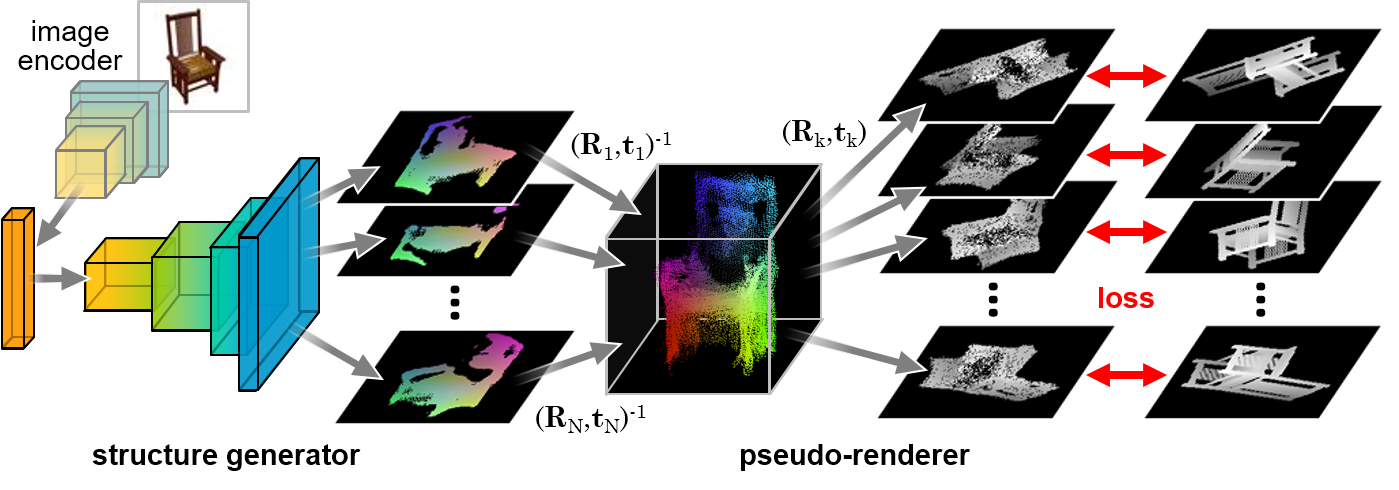}
\caption{\textbf{Network architecture.} 
From an encoded latent representation, we propose to use a structure generator (Sec~\ref{sec:gen}), which is based on \textit{2D convolutional} operations, to predict the 3D structure at $N$ viewpoints.
The point clouds are fused by transforming the 3D structure at each viewpoint to the canonical coordinates.
The pseudo-renderer (Sec.~\ref{sec:loss}) synthesizes depth images from novel viewpoints, which are further used for joint 2D projection optimization.
This contains no learnable parameters and reasons based purely on 3D geometry.}
\label{fig:network}
\end{figure*}

Our goal is to generate 3D predictions that compactly shape the surface geometry with dense point clouds.
The overall pipeline is illustrated in Fig.~\ref{fig:network}.
We start with an encoder that maps the input data to a latent representation space.
The encoder may take on various forms of data depending on the application; in our experiments, we focus on encoding RGB images for single-image 3D reconstruction tasks.
From the latent representation, we propose to generate the dense point clouds using a structure generator based on 2D convolutions with a joint 2D projection criterion, described in detail as follows.

\subsection{Structure Generator} \label{sec:gen}

The structure generator predicts the 3D structure of the object at $N$ different viewpoints (along with their binary masks), \ie the 3D coordinates $\hat{\x}_i = [\hat{x}_i \; \hat{y}_i \; \hat{z}_i]^{\top}$ at each pixel location.
Pixel values in natural images can be synthesized through convolutional generative models mainly due to their exhibition of strong local spatial dependencies; similar phenomenons can be observed for point clouds when treating them as $(x,y,z)$ multi-channel images on a 2D grid.
Based on this insight, the structure generator is mainly based on 2D convolutional operations to predict the $(x,y,z)$ images representing the 3D surface geometry.
This approach circumvents the need of time-consuming and memory-expensive 3D convolutional operations for volumetric predictions.
The evidence of such validity is verified in our experimental results.

Assuming the 3D rigid transformation matrices of the $N$ viewpoints $(\R_1,\t_1) ... (\R_N,\t_N)$ are given a priori, each 3D point $\hat{\x}_i$ at viewpoint $n$ can be transformed to the canonical 3D coordinates as $\hat{\p}_i$ via
\begin{equation} \label{eq:fuse}
\hat{\p}_i = \R_n^{-1} \left( \K^{-1} \hat{\x}_i - \t_n \right)  \;\;\; \forall i \; ,
\end{equation}
where $\K$ is the predefined camera intrinsic matrix.
This defines the relationship between the predicted 3D points and the fused collection of point clouds in the canonical 3D coordinates, which is the outcome of our network.

\subsection{Joint 2D Projection Optimization} \label{sec:loss}

To learn point cloud generation using the provided 3D CAD models as supervision, the standard approach would be to optimize over a 3D-based metric that defines the distance between the point cloud and the ground-truth CAD model (\eg Chamfer distance~\cite{fan2016point}).
Such metric usually involves computing surface projections for every generated point, which can be computationally expensive for very dense predictions, making it intractable.

We overcome this issue by alternatively optimizing over the \textit{joint 2D projection error} of novel viewpoints.
Instead of using only projected binary masks as supervision~\cite{yan2016perspective,gadelha20163d,rezende2016unsupervised}, we conjecture that a well-generated 3D shape should also have the ability to render reasonable depth images from any viewpoint.
To realize this concept, we introduce the pseudo-renderer, a differentiable module to approximate true rendering, to synthesize novel depth images from dense point clouds.

\paragraph{Pseudo-rendering.}
Given the 3D rigid transformation matrix of a novel viewpoint $(\R_k,\t_k)$, each canonical 3D point $\hat{\p}_i$ can be further transformed to $\hat{\x}'_i$ back in the image coordinates via
\begin{equation} \label{eq:render}
\hat{\x}'_i = \K \left( \R_k \hat{\p}_i + \t_k \right) \;\;\; \forall i \; .
\end{equation}
This is the inverse operation of Eq.~\eqref{eq:fuse} with different transformation matrices and can be combined with Eq.~\eqref{eq:fuse} together, composing a single effective transformation.
By such, we obtain the $(\hat{x}'_i,\hat{y}'_i)$ location as well as the new depth value $\hat{z}'_i$ at viewpoint $k$.

\begin{figure}[t!] \center
\includegraphics[width=0.95\linewidth]{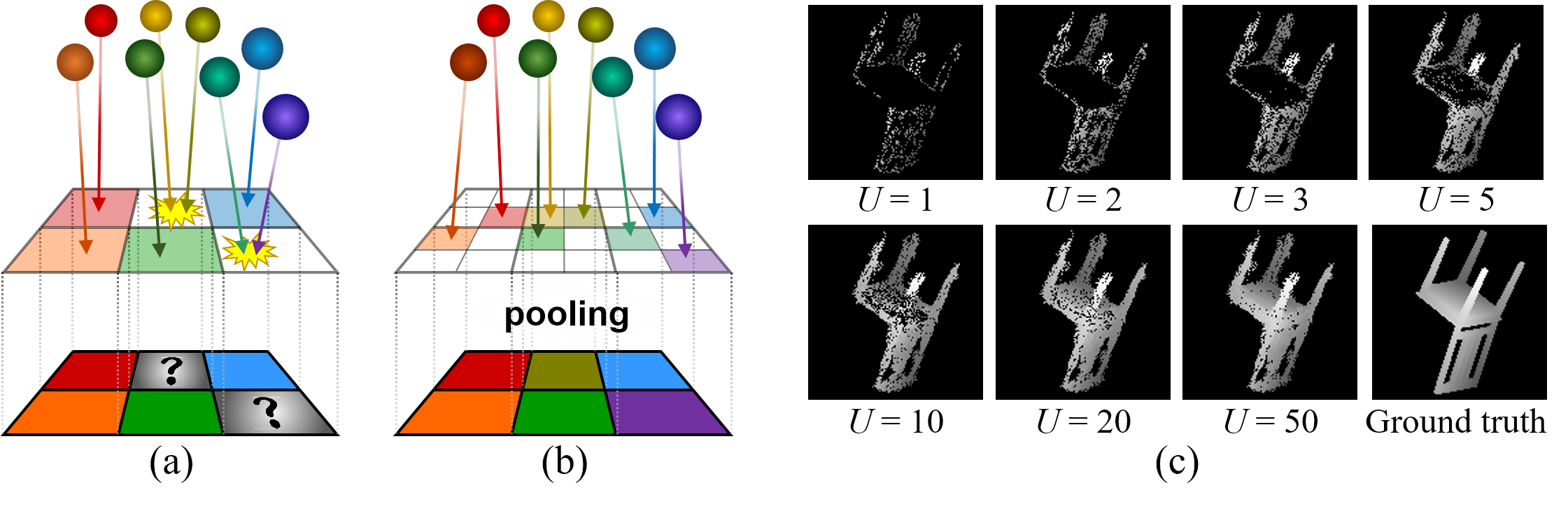}
\caption{\textbf{Concept of pseudo-rendering.} Multiple transformed 3D points may correspond to projection on the same pixels in the image space.
(a) Collision could easily occur if $(\hat{x}'_i,\hat{y}'_i)$ were directly discretized.
(b) Upsampling the target image increases the precision of the projection locations and thus alleviates the collision effect.
A max-pooling operation on the inverse depth values follows as to obtain the original resolution while maintaining the effective depth value at each pixel.
(c) Examples of pseudo-rendered depth images with various upsampling factors $U$ (only valid depth values without collision are shown).
Pseudo-rendering achieves closer performance to true rendering with a higher value of $U$.}
\label{fig:render}
\end{figure}

To produce a pixelated depth image, one would also need to discretize all $(\hat{x}'_i,\hat{y}'_i)$ coordinates, resulting in possibly multiple transformed points projecting and ``colliding'' onto the same pixel (Fig.~\ref{fig:render}).
We resolve this issue with the pseudo-renderer $f_\text{PR}(\cdot)$, which increases the projection resolution to alleviate such collision effect.
Specifically, $\hat{\x}'_i$ is projected onto a target image upsampled by a factor of $U$, reducing the quantization error of $(\hat{x}'_i,\hat{y}'_i)$ as well as the probability of collision occurrence.
A max-pooling operation on the inverse depth values with kernel size $U$ follows to downsample back to the original resolution while maintaining the minimum depth value at each pixel location.
We use such approximation of the rendering operation to maintain differentiability and parallelizability within the backpropagation framework.

\paragraph{Optimization.}
We use the pseudo-rendered depth images $\hat{\Z} = f_\text{PR} (\{\hat{\x}'_i\})$ and the resulting masks $\hat{\M}$ at novel viewpoints for optimization.
The loss function consists of the mask loss $\L_\text{mask}$ and the depth loss $\L_\text{depth}$, respectively defined as
\begin{align} \label{eq:2dloss-2}
\L_\text{mask} = \sum_{k=1}^K - \M_k \log \hat{\M}_k - \left( 1-\M_k \right) \log \left( 1-\hat{\M}_k \right) \;\;\; \text{and} \;\;\;
\L_\text{depth} = \sum_{k=1}^K \left\| \hat{\Z}_k - \Z_k \right\|_1 \;\; ,
\end{align}
where we simultaneously optimize over $K$ novel viewpoints at a time.
$\M_k$ and $\Z_k$ are the ground-truth mask and depth images at the $k$th novel viewpoint.
We use element-wise $L_1$ loss for the depth (posing it as a pixel-wise binary classification problem) and cross-entropy loss for the mask.
The overall loss function is defined as $\L = \L_\text{mask} + \lambda \cdot \L_\text{depth}$, where $\lambda$ is the weighting factor.

Optimizing the structure generator over novel projections enforces joint 3D geometric reasoning between the predicted point clouds from the $N$ viewpoints.
It also allows the optimization error to evenly distribute across novel viewpoints instead of focusing on the fixed $N$ viewpoints.

\section{Experiments}

We evaluate our proposed method by analyzing its performance in the application of single-image 3D reconstruction and comparing against state-of-the-art methods.

\paragraph{Data preparation.}
We train and evaluate all networks using the ShapeNet database~\cite{chang2015shapenet}, which contains a large collection of categorized 3D CAD models. 
For each CAD model, we pre-render 100 depth/mask image pairs of size 128$\times$128 at random novel viewpoints as the ground truth of the loss function.
We consider the entire space of possible 3D rotations (including in-plane rotation) for the viewpoints and assume identity translation for simplicity.
The input images are objects pre-rendered from a fixed elevation and 24 different azimuth angles.

\begin{table}[t!] \center
\begin{tabular}{c c c l l}
\hline
\multirow{2}{*}{\bf Sec.} & \bf Input & \bf Latent & \multicolumn{2}{c}{\bf Number of filters} \\
& \bf size & \bf vector & \multicolumn{1}{c}{image encoder} & \multicolumn{1}{c}{structure generator} \\ \hline \hline
\multirow{2}{*}{\ref{sec:single}} & \multirow{2}{*}{64$\times$64} & \multirow{2}{*}{512-D} & conv: 96, 128, 192, 256 & linear: 1024, 2048, 4096 \\
& & & linear: 2048, 1024, 512 & deconv: 192, 128, 96, 64, 48 \\ \hline
\multirow{2}{*}{\ref{sec:multi}} & \multirow{2}{*}{128$\times$128} & \multirow{2}{*}{1024-D} & conv: 128, 192, 256, 384, 512 & linear: 2048, 4096, 12800 \\
& & & linear: 4096, 2048, 1024 & deconv: 384, 256, 192, 128, 96 \\ \hline
\end{tabular}
\caption{\textbf{Architectural details} of our proposed method for each experiment.}
\label{table:architecture}
\end{table}

\paragraph{Architectural details.}
The structure generator follows the structure of conventional deep generative models, consisting of linear layers followed by 2D convolution layers (with kernel size 3$\times$3).
The dimensions of all feature maps are halved after each encoder convolution and doubled after each generator convolution.
Details of network dimensions are listed in Table~\ref{table:architecture}.
At the end of the decoder, we add an extra convolution layer with filters of size 1$\times$1 to encourage individuality of the generated pixels.
Batch normalization~\cite{ioffe2015batch} and ReLU activations are added between all layers.

The generator predicts $N=8$ images of size 128$\times$128 with 4 channels ($x$, $y$, $z$ and the binary mask), where the fixed viewpoints are chosen from the 8 corners of a centered cube.
Orthographic projection is assumed in the transformation in~\eqref{eq:fuse} and~\eqref{eq:render}.
We use $U=5$ for the upsampling factor of the pseudo-renderer in our experiments.

\paragraph{Training details.}
All networks are optimized using the Adam optimizer~\cite{kingma2014adam}.
We take a two-stage training procedure: the structure generator is first pretrained to predict the depth images from the $N$ viewpoints (with a constant learning rate of 1e-2), and then the entire network is fine-tuned with joint 2D projection optimization (with a constant learning rate of 1e-4).
For the training parameters, we set $\lambda=1.0$ and $K=5$.

\paragraph{Quantitative metrics.}
We measure using the average point-wise 3D Euclidean distance between two 3D models: for each point $\hat{\p}_i$ in the source model, the distance to the target model $\mathcal{S}$ is defined as $\mathcal{E}_i = \min_{\p_j \in \mathcal{S}} \left\| \hat{\p}_i - \p_j  \right\|_2$.
This metric is defined bidirectionally as the distance from the predicted point cloud to the ground-truth CAD model and vice versa. It is necessary to report both metrics for they represent different aspects of quality --- the former measures 3D shape similarity and the latter measures surface coverage~\cite{kong2017using}.
We represent the ground-truth CAD models as collections of uniformly densified 3D points on the surfaces (100K densified points in our settings).

\subsection{Single Object Category} \label{sec:single}

We start by evaluating the efficacy of our dense point cloud representation on 3D reconstruction for a single object category.
We use the chair category from ShapeNet, which consists of 6,778 CAD models.
We compare against (a) Tatarchenko~\etal~\cite{tatarchenko2016multi}, which learns implicit 3D representations through a mixed embedding, and (b) Perspective Transformer Networks (PTN)~\cite{yan2016perspective}, which learns to predict volumetric data by minimizing the projection error.
We include two variants of PTN as well as a baseline 3D ConvNet from Yan~\etal~\cite{yan2016perspective}.
We use the same 80\%-20\% training/test split provided by Yan~\etal~\cite{yan2016perspective}.

We pretrain our network for 200K iterations and fine-tune end-to-end for 100K iterations.
For the method of Tatarchenko~\etal~\cite{tatarchenko2016multi}, we evaluate by predicting depth images from our same $N$ viewpoints and transform the resulting point clouds to the canonical coordinates.
This shares the same network architecture to ours, but with 3D pose information additionally encoded using 3 linear layers (with 64 filters) and concatenated with the latent vector.
We use the novel depth/mask pairs as direct supervision for the decoder output and train this network for 300K iterations with a constant learning rate of 1e-2.
For PTN~\cite{yan2016perspective}, we extract the surface voxels (by subtracting the prediction by its eroded version) and rescale them such that the tightest 3D bounding boxes of the prediction and the ground-truth CAD models have the same volume.
We use the pretrained models readily provided by the authors.
%

\begin{table}[t!]
\begin{tabular}{l c c}
\hline
\multicolumn{1}{c}{\multirow{2}{*}{\bf Method}} & \multicolumn{2}{c}{\bf 3D error metric} \\
& pred. $\to$ GT & GT $\to$ pred. \\ \hline \hline
3D ConvNet (vol. only)~\cite{yan2016perspective} & 1.827 & 2.660 \\
PTN (proj. only)~\cite{yan2016perspective} & 2.181 & 2.170 \\
PTN (vol. \& proj.)~\cite{yan2016perspective} & 1.840 & 2.585 \\
Tatarchenko~\etal~\cite{tatarchenko2016multi} & 2.381 & 3.019 \\
Proposed method & \bf 1.768 & \bf 1.763 \\ \hline
\end{tabular}
\caption{\textbf{Average 3D test error} of the single-category experiment.
Our method outperforms all baselines in both metrics, indicating the superiority in fine-grained shape similarity and point cloud coverage on the surface.
(All numbers are scaled by 0.01)
\label{table:single}}
\end{table}

The quantitative results on the test split are reported in Table~\ref{table:single}.
We achieve a lower average 3D distance than all baselines in both metrics, even though our approach is optimized with joint 2D projections instead of these 3D error metrics.
This demonstrates that we are capable of predicting more accurate shapes with higher density and finer granularity.
This highlights the efficiency of our approach using 2D ConvNets to generate 3D shapes compared to 3D ConvNet methods such as PTN~\cite{yan2016perspective} as they attempt to predict all voxel occupancies inside a 3D grid space.
Compared to Tatarchenko~\etal~\cite{tatarchenko2016multi}, an important takeaway is that 3D geometry should explicitly factorized when possible instead of being implicitly learned by the network parameters.
It is much more efficient to focus on predicting the geometry from a sufficient number of viewpoints and combining them with known geometric transformations.

\begin{figure}[t!]
\includegraphics[width=\linewidth]{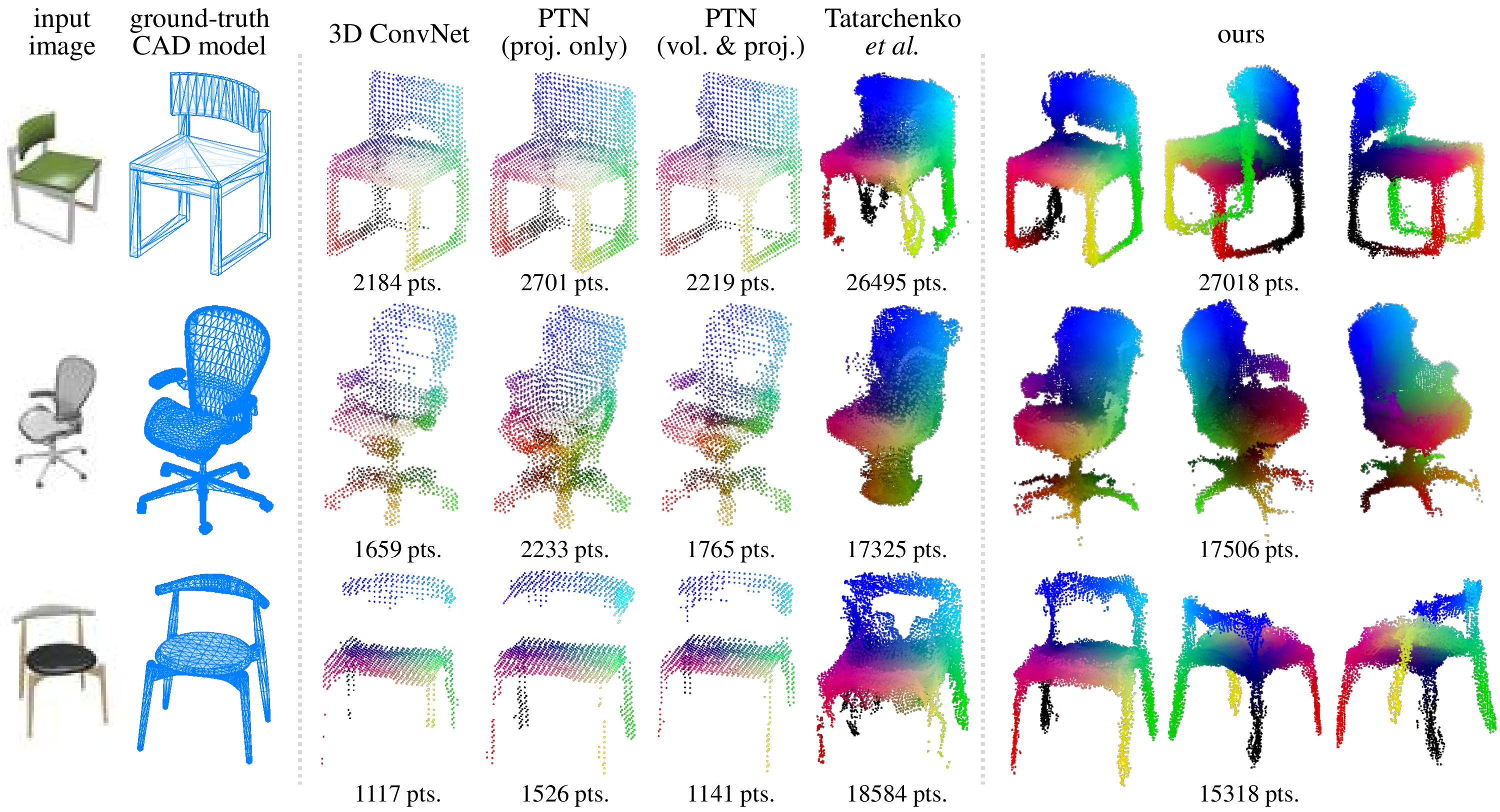}
\caption{\textbf{Qualitative results} from the single-category experiment.
Our method generates denser predictions compared to the volumetric baselines and more accurate shapes than Tatarchenko~\etal~\cite{tatarchenko2016multi}, which learns 3D synthesis implicitly.
The RGB values of the point cloud represents the 3D coordinate values.
Best viewed in color.
}
\label{fig:single}
\end{figure}

We visualize the generated 3D shapes in Fig.~\ref{fig:single}.
Compared to the baselines, we predict more accurate object structures with a much higher point cloud density (around 10$\times$ higher than $32^3$ volumetric methods).
This further highlights the desirability of our approach --- we are able to efficiently use 2D convolutional operations and utilize high-resolution supervision given similar memory budgets.

\subsection{General Object Categories} \label{sec:multi}

We also evaluate our network on the single-image 3D reconstruction task trained with multiple object categories.
We compare against (a) 3D-R2N2~\cite{choy20163d}, which learns volumeric predictions through recurrent networks, and (b) Fan~\etal~\cite{fan2016point}, which predicts an unordered set of 1024 3D points.
We use 13 categories of ShapeNet for evaluation (listed in Table~\ref{table:multi}), where the 80\%-20\% training/test split is provided by Choy~\etal~\cite{choy20163d}.
We evaluate 3D-R2N2 by its surface voxels using the same procedure as described in Sec.~\ref{sec:single}.
We pretrain our network for 300K iterations and fine-tune end-to-end for 100K iterations; for the baselines, we use the pretrained models readily provided by the authors.

\begin{table}[t!]
\begin{tabular}{c||c|c|c|c|c}
\hline
\multicolumn{1}{c||}{\multirow{2}{*}{\bf Category}} & \multicolumn{3}{c|}{3D-R2N2~\cite{choy20163d}} & Fan~\etal~\cite{fan2016point} & Proposed \\
& 1 view    & 3 views & 5 views & (1 view) & (1 view) \\ \hline
airplane & 3.207 / 2.879 & 2.521 / 2.468 & 2.399 / 2.391 & 1.301 / {\bf 1.488} & {\bf 1.294} / 1.541 \\
bench & 3.350 / 3.697 & 2.465 / 2.746 & 2.323 / 2.603 & 1.814 / 1.983 & {\bf 1.757} / {\bf 1.487} \\
cabinet & 1.636 / 2.817 & 1.445 / 2.626 & {\bf 1.420} / 2.619 & 2.463 / 2.444 & 1.814 / {\bf 1.072} \\
car & 1.808 / 3.238 & 1.685 / 3.151 & 1.664 / 3.146 & 1.800 / 2.053 & {\bf 1.446} / {\bf 1.061} \\
chair & 2.759 / 4.207 & 1.960 / 3.238 & {\bf 1.854} / 3.080 & 1.887 / 2.355 & 1.886 / {\bf 2.041} \\
display & 3.235 / 4.283 & 2.262 / 3.151 & 2.088 / 2.953 & {\bf 1.919} / 2.334 & 2.142 / {\bf 1.440} \\
lamp & 8.400 / 9.722 & 6.001 / 7.755 & 5.698 / 7.331 & {\bf 2.347} / {\bf 2.212} & 2.635 / 4.459 \\
loudspeaker & 2.652 / 4.335 & 2.577 / 4.302 & 2.487 / 4.203 & 3.215 / 2.788 & {\bf 2.371} / {\bf 1.706} \\
rifle & 4.798 / 2.996 & 4.307 / 2.546 & 4.193 / 2.447 & 1.316 / {\bf 1.358} & {\bf 1.289} / 1.510 \\
sofa & 2.725 / 3.628 & 2.371 / 3.252 & 2.306 / 3.196 & 2.592 / 2.784 & {\bf 1.917} / {\bf 1.423} \\
table & 3.118 / 4.208 & 2.268 / 3.277 & 2.128 / 3.134 & 1.874 / 2.229 & {\bf 1.689} / {\bf 1.620} \\
telephone & 2.202 / 3.314 & 1.969 / 2.834 & 1.874 / 2.734 & {\bf 1.516} / 1.989 & 1.939 / {\bf 1.198} \\
watercraft & 3.592 / 4.007 & 3.299 / 3.698 & 3.210 / 3.614 & {\bf 1.715} / 1.877 & 1.813 / {\bf 1.550} \\
\hline
 \bf mean& 3.345 / 4.102 & 2.702 / 3.465 & 2.588 / 3.342 & 1.982 / 2.146 & {\bf 1.846} / {\bf 1.701} \\
\hline
\end{tabular}
\caption{\textbf{Average 3D test error} of the multi-category experiment, where the numbers are shown as [\;prediction$\to$GT / GT$\to$prediction\;].
The mean is computed across categories.
For the single-view case, we outperform all baselines in 8 and 10 out of 13 categories for the two 3D error metrics.
(All numbers are scaled by 0.01)}
\label{table:multi}
\end{table}

\begin{figure}[t!]
\includegraphics[width=\linewidth]{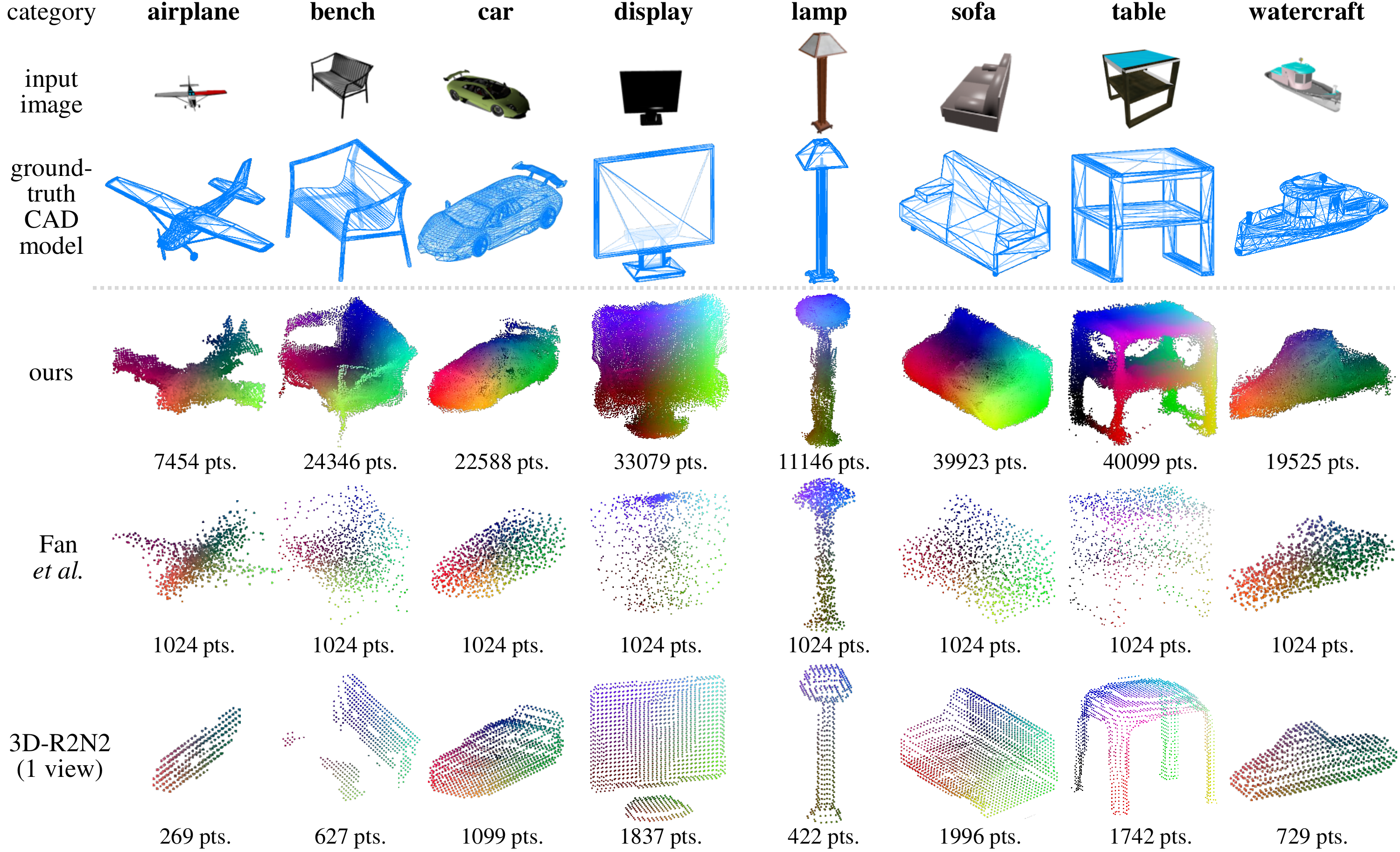}
\caption{\textbf{Qualitative results} from the multi-category experiment.
Our method generates denser and more certain predictions compared to the baselines.
}
\label{fig:multi}
\end{figure}

We list the quantitative results in Table~\ref{table:multi}, where the metrics are reported per-category.
Our method achieves an overall lower error in both metrics.
We outperform the volumetric baselines (3D-R2N2) by a large margin and has better prediction performance than Fan~\etal in most cases.
We also visualize the predictions in Fig.~\ref{fig:multi}; again we see that our method predicts more accurate shapes with higher point density.
We find that our method can be more problematic when objects contain very thin structures (\eg lamps); adding hybrid linear layers~\cite{fan2016point} may help improve performance.

\subsection{Generative Representation Analysis} \label{sec:analysis}

We analyze the learned generative representations by observing the 3D predictions from manipulation in the latent space.
Previous works have demonstrated that deep generative networks can generate meaningful pixel/voxel predictions by performing linear operations in the latent space~\cite{radford2015unsupervised,dosovitskiy2015learning,wu2016learning}; here, we explore the possibility of such manipulation for dense point clouds in an undiscretized space.

\begin{figure}[t!]
\includegraphics[width=\linewidth]{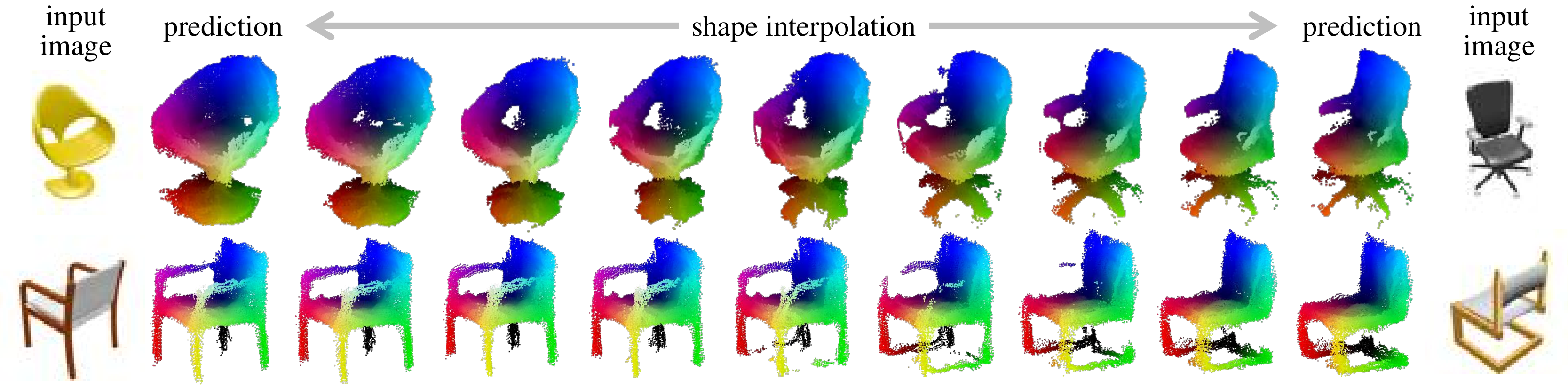}
\caption{Dense shapes generated from interpolated latent embeddings of two input images (leftmost and rightmost).
The interpolated shapes maintain reasonable structures of chairs.}
\vspace{-4pt}
\label{fig:interp}
\end{figure}

\begin{figure}[t!]
\includegraphics[width=\linewidth]{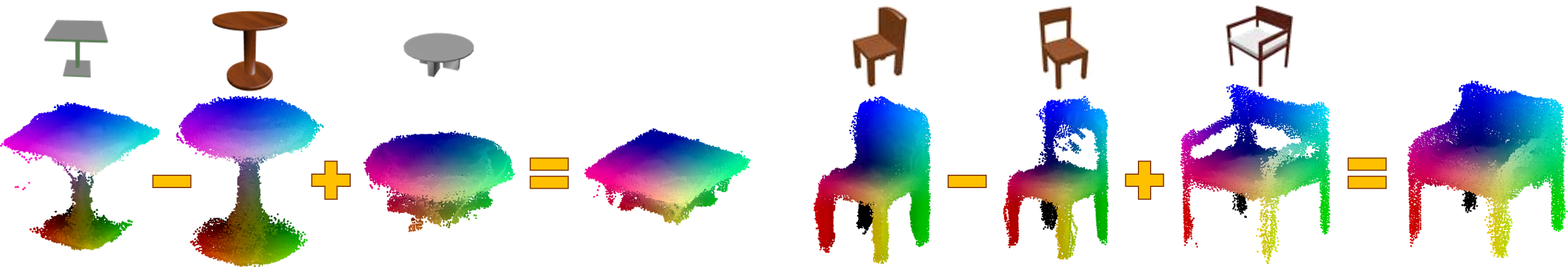}
\caption{Dense shapes generated from arithmetic operations in the latent space (left: tables, right: chairs), where the input images are shown in the top row.
}
\label{fig:arith}
\end{figure}

We show in Fig.~\ref{fig:interp} the resulting dense shapes generated from the embedding vector interpolated in the latent space.
The morphing transition is smooth with plausible interpolated shapes, which suggests that our structure generator can generate meaningful 3D predictions from convex combinations of encoded latent vectors.
The structure generator is also capable of generating reasonable novel shapes from arithmetic results in the latent space --- from Fig.~\ref{fig:arith}) we observe semantic feature replacement of table height/shape as well as chair arms/backs.
These results suggest that the high-level semantic information encoded in the latent vectors are manipulable and interpretable of the resulting dense point clouds through the structure generator.

\section{Conclusion}
In this paper, we introduced a framework for generating 3D shapes in the form of dense point clouds.
Compared to conventional volumetric prediction methods using 3D ConvNets, it is more efficient to utilize 2D convolutional operations to predict surface information of 3D shapes.
We showed that by introducing a pseudo-renderer, we are able to synthesize approximate depth images from novel viewpoints to optimize the 2D projection error within a backpropagation framework.
Experimental results for single-image 3D reconstruction tasks showed that we generate more accurate and much denser 3D shapes than state-of-the-art 3D reconstruction methods.


{\small
\bibliographystyle{ieee}
\bibliography{reference}
}

%
%
%

\end{document}